\PassOptionsToPackage{numbers,sort&compress}{natbib}
\documentclass[sigconf,nonacm]{acmart}

\usepackage[utf8]{inputenc}
\usepackage[T1]{fontenc}
\usepackage{microtype}
\usepackage{booktabs}
\usepackage{tabularx}
\usepackage{multirow}
\usepackage{float}
\usepackage{amsmath,amssymb}
\usepackage{xcolor}
\usepackage{tikz}
\usepackage{url}
\usetikzlibrary{arrows.meta,positioning,fit}

\setcopyright{none}
\hypersetup{
  hidelinks,
  pdftitle={ClaimReceipt: Verifying Evidence Sufficiency and Coverage in Agent Evaluations},
  pdfauthor={Peiying Zhu; Sidi Chang}
}

\definecolor{passgreen}{HTML}{2E7D5B}
\definecolor{invalidred}{HTML}{B54747}
\definecolor{incamber}{HTML}{B7791F}
\definecolor{receiptblue}{HTML}{2B6F9F}
\definecolor{softgray}{HTML}{EFF2F4}

\newcommand{\pass}{\textsc{Pass}}
\newcommand{\invalid}{\textsc{Invalid}}
\newcommand{\inconclusive}{\textsc{Inconclusive}}
\newcommand{\claimreceipt}{\textsc{ClaimReceipt}}

\begin{document}
\acmshorttitle{ClaimReceipt}
\acmshortauthors{Zhu \& Chang}
\pagestyle{acmstyle}
\thispagestyle{firstpage}

\twocolumn[
\begin{@twocolumnfalse}
\begin{center}
{\LARGE\bfseries \claimreceipt: Verifying Evidence Sufficiency and Coverage in Agent Evaluations\par}
\vspace{1.0em}

\begin{tabular}{@{}c@{\hspace{2em}}c@{}}
{\large Peiying Zhu\textsuperscript{*}} & {\large Sidi Chang\textsuperscript{*\,$\dagger$}} \tabularnewline
{\small peiying@blossomai.co} & {\small schang@blossomai.co} \tabularnewline
{\small Blossom AI} & {\small Blossom AI Labs} \tabularnewline
{\small San Francisco, CA, USA} & {\small Tokyo, Japan}
\end{tabular}
\vspace{1.2em}
\end{center}

\noindent\parbox{\textwidth}{
\small
\noindent\textbf{Abstract}\\[3pt]
Agent evaluations face two distinct evidentiary questions: whether a reported claim is recomputable from retained evidence (\emph{sufficiency}), and whether the retained records cover the committed experiment set (\emph{coverage}). Generic logs and hash-linked transcripts answer neither reliably. We introduce \claimreceipt, a claim-relative receipt specification and selective verifier that binds typed transaction evidence to a signed experiment manifest and returns \pass, \invalid, or \inconclusive{} per claim. We freeze the specification before implementation (SHA-256 \texttt{18d109...b81}). On 1,392 historical buyer--seller records, a CR-2 verifier reproduces all five manually labeled audit verdicts, exactly replays 600 deterministic and 792 post-generation records, makes every one of 13 declared field groups non-redundant under tested ablations, and returns the expected result on 11/11 semantic faults with 0/8 false positives. We then run a separate prospective CR-3 epoch: 30 assignments are committed before inference, terminal receipts are signed and chained, and private evidence is encrypted for an auditor. Complete evidence yields coverage and accounting \pass{}; withholding one terminal receipt returns \texttt{INCONCLUSIVE\_COVERAGE}, while withholding all private openings preserves coverage and protocol verification but makes economic claims inconclusive, exactly matching a preregistered prediction. Receipt instrumentation adds 0.021\% of model-inference time and 9.9\,KB per transaction. A specification-legibility probe indicates that our own frozen specification is not yet unambiguous to an independent reader. Claim verification therefore requires both claim-sufficient evidence and a committed universe against which omissions become visible.
}

\vspace{1.5em}
\end{@twocolumnfalse}
]

\renewcommand{\thefootnote}{\fnsymbol{footnote}}
\footnotetext[1]{Both authors contributed equally to this research.}
\footnotetext[2]{Corresponding author.}
\renewcommand{\thefootnote}{\arabic{footnote}}
\setcounter{footnote}{0}

\section{Introduction}

When an agent benchmark reports that a prompt, scaffold, guardrail, or model improved performance, two questions follow. \emph{Sufficiency} asks whether the number can be recomputed correctly from retained evidence. \emph{Coverage} asks whether those records are the complete committed set rather than a favorable subset. Sufficiency protects against an honest but fallible evaluator; coverage also addresses an operator that may prefer some runs to remain unseen. A final reward, transcript, or generic signed log may preserve bytes while omitting private state, parser outputs, discarded attempts, or an expected terminal record.

This gap is especially acute for interactive agent systems. Their measured outcome is jointly produced by model generations, hidden environment state, parsers, enforcement code, retry rules, and downstream scoring. Agent benchmarks already face large run-to-run variance and protocol sensitivity \cite{kapoor2025agents,liang2022helm}. In multi-agent settings, the environment also determines what one agent can learn and which actions another may take. A scalar can therefore remain numerically correct while the causal interpretation attached to it is invalid.

Prior work audited this problem in an agent-to-agent (A2A) commerce simulation \cite{zhu2026construct}. A manual construct-validity contract separated four questions: whether the seller instantiated the intended incentive (C1), whether treatment arms differed only in the intervention (C2), whether stochastic estimates were stable (C3), and whether outcome accounting was complete (C4). That audit stopped an attractive policy claim: the original experiment was \invalid{} because the arms used different offer protocols, while a controlled rerun remained \inconclusive{} because incentive validity and generation stability did not pass. The audit also established an identifiability boundary: if an evidence abstraction maps executions with different claim truth to the same retained record, no downstream metric can recover the discarded distinction.

This paper gives the constructive counterpart. \claimreceipt{} asks what a platform must retain so a bounded claim is recomputable, and what it must commit so missing runs are detectable. Hashing everything is not enough. The contribution is a typed evidence schema, an experiment-level manifest, exact accounting semantics, and a selective decision procedure. Hashes, signatures, and tamper-evident chains are standard transport mechanisms \cite{haber1991timestamp,laurie2013certificate}; they cannot decide which semantic distinctions matter or define the universe that should have produced receipts.

\begin{figure*}[t]
\centering
\begin{tikzpicture}[
  node distance=4mm and 4mm,
  box/.style={draw,rounded corners=1pt,minimum height=7mm,align=center,font=\scriptsize,inner xsep=4pt},
  arrow/.style={-{Latex[length=2mm]},thick},
]
\node[box,fill=softgray] (manifest) {prospective manifest\\assignments, protocol, estimator};
\node[box,fill=softgray,right=of manifest] (run) {agent transaction\\trace, offers, attempts};
\node[box,fill=receiptblue!12,right=of run] (receipt) {claim receipt\\public + auditor evidence};
\node[box,fill=passgreen!12,right=of receipt] (verify) {selective verifier\\\pass{} / \invalid{} / \inconclusive{}};
\draw[arrow] (manifest) -- (run);
\draw[arrow] (run) -- (receipt);
\draw[arrow] (receipt) -- (verify);
\draw[arrow,bend left=28] (manifest.north east) to node[above,font=\scriptsize]{digest and assignment contract} (receipt.north west);
\end{tikzpicture}
\caption{\claimreceipt{} binds transaction evidence to an experiment manifest. Transport integrity protects retained bytes; claim semantics determine what those bytes license.}
\label{fig:architecture}
\end{figure*}
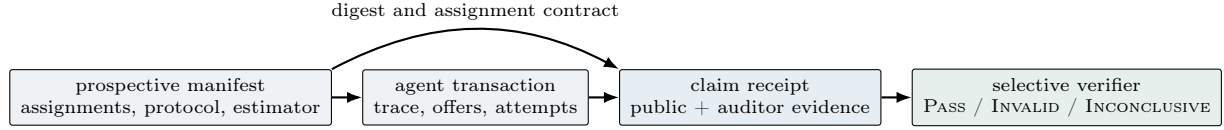

Figure~\ref{fig:architecture} shows the design. Before prospective ingress, a manifest commits the population, arm definitions, protocol fingerprints, replication structure, estimator, thresholds, and coverage policy. Each transaction receipt links that manifest to an assignment, private-profile commitment, ordered trace, raw and normalized offers, chooser decision, exact economic claims, and artifact digests. The reference verifier checks integrity and coverage, replays claims from the appropriate boundary, and returns a per-contract verdict. It may license descriptive reporting while blocking a causal policy claim.

Our contributions are:
\begin{enumerate}
  \item We formalize claim-relative evidence sufficiency over executions, define a bounded claim class, and provide a field-to-claim dependency matrix. The schema is sufficient by construction for that class, not for arbitrary claims about intent or external truth.
  \item The verifier performs selective verification at CR-2 and CR-3: it recomputes transaction and batch claims, preserves \inconclusive{} as a first-class output, and adds prospective manifest, ingress, encrypted-opening, receipt-chain, and conditional-coverage checks.
  \item On a previously audited corpus of 1,392 receipts across six epochs, the verifier agrees with 5/5 manual verdicts, exactly replays every declared transaction claim, and reports the complete historical $4\times6$ grid plus a separate prospective row, rather than only labeled cells.
  \item Field ablations, 11 frozen faults, and eight benign variations test schema semantics; a separate 30-transaction prospective evaluation tests complete, missing-terminal, and missing-private-evidence cases with measured overhead.
\end{enumerate}

\section{Problem Formulation}

\subsection{Claim-relative sufficiency}

Let $\tau$ be a complete admissible execution, $c(\tau)$ the ground-truth value of claim $c$, and $\phi_E(\tau)$ the evidence retained by field set $E$. We call $E$ \emph{claim-sufficient} for $c$ exactly when
\begin{equation}
\forall \tau_1,\tau_2:\quad \phi_E(\tau_1)=\phi_E(\tau_2) \Rightarrow c(\tau_1)=c(\tau_2).
\label{eq:sufficiency}
\end{equation}
This definition is over executions and claim truth, not a verifier's output. A verifier that always abstains cannot make $E=\emptyset$ sufficient. Conversely, when required evidence is absent, a verifier must not substitute a reported scalar; it must reject or abstain.

Sufficiency is always relative to a claim class. Version 0.3 supports eight transaction claims---completion, buyer surplus (BS), seller profit (SP), welfare (W), first-best fraction, match quality, and two treatment-fidelity claims---plus four experiment claims C1--C4. It does not claim sufficiency for latent intent, equilibrium, fairness, external profile truth, or uncommitted transactions.

Coverage is instead a property of a set: no single receipt can show that every committed assignment produced a terminal record. It requires an experiment manifest and an ingress commitment that define the expected set before outcomes are observed.

\subsection{Running transaction semantics}

We instantiate the specification in configurable buyer--seller transactions. A buyer profile $\theta$ assigns private values to a component set $S$, an outside-option utility $o_\theta$, and hard constraints. If an offer at price $p$ is accepted,
\begin{align}
\mathrm{BS}_\theta(S,p)&=V_\theta(S)-p-o_\theta,\\
\mathrm{SP}(S,p)&=p-C(S),\\
W_\theta(S)&=V_\theta(S)-C(S)-o_\theta=\mathrm{BS}+\mathrm{SP}.
\end{align}
Rejected or failed transactions contribute zero. First-best welfare is $W^\star_\theta=\max(0,\max_S W_\theta(S))$ under the committed component rule. The first-best claim is the exact rational pair $(W,W^\star)$ when $W^\star>0$ and a typed null otherwise; it is not clamped, because negative welfare records destruction. Match quality is also retained as an exact numerator--denominator pair.

The experiment claims encode the prior audit contract \cite{zhu2026construct}: C1 tests whether ordered seller-objective levels produce the declared directional response; C2 tests cross-arm protocol isolation; C3 tests stochastic stability using profile-level pairing with replications nested inside profile--condition cells; and C4 tests replay and complete accounting. Each returns \pass, \invalid, or \inconclusive{}.

\section{ClaimReceipt Design}

\subsection{Four evidence layers}

The specification separates four layers that generic ``audit log'' language often conflates.
\begin{enumerate}
  \item Transport integrity covers exact bytes, sequence numbers, chain links, and signatures. It detects mutation of evidence already committed.
  \item Semantic evidence (sufficiency) retains typed profile, trace, offer-stage, choice, catalog, and scorer inputs needed to recompute claims.
  \item Experiment coverage binds receipts to an expected assignment matrix and ingress commitment, making omissions after committed ingress visible.
  \item External truth remains outside the system. A platform cannot prove that an unobserved interaction occurred or that a synthetic profile describes a real person.
\end{enumerate}
The distinction prevents cryptography from being mistaken for construct validity. Certificate Transparency and tamper-evident logs establish valuable integrity properties \cite{laurie2013certificate,crosby2009efficient}; they do not decide whether an omitted private profile makes welfare underdetermined.

\subsection{Threat model and trust boundary}

Transport mechanisms address L1 byte mutation; semantic replay and protocol fingerprints address L2 construct drift; prospective ingress and terminal reconciliation address L3 omissions after commitment. L4 remains open: no receipt system can detect transactions prevented from reaching the witness or a false but internally consistent ground-truth profile.

\begin{table}[t]
\caption{Prototype trust roles. Separation is by key and function, not by organizational principal: one operator controls all three local roles.}
\label{tab:threat}
\centering
\scriptsize
\setlength{\tabcolsep}{3pt}
\begin{tabularx}{\linewidth}{lXX}
\toprule
Role & Trusted action & Residual capability / boundary \\
\midrule
Manifest authority & Ed25519-sign contract before ingress & Can declare a biased universe or false profile (L4) \\
Ingress authority & Ed25519-sign tickets, checkpoint, receipts & Separate role, but may collude with owner in this prototype \\
Runner & Produce terminal evidence after a ticket & May mutate or omit; chains and reconciliation expose L1/L3 \\
Auditor & Hold X25519 opening key and replay private evidence & Learns sensitive profiles; governance and deletion remain external \\
OTS calendars & Commit manifest digest before first ticket & External time anchor, not an independent ingress witness \\
\bottomrule
\end{tabularx}
\end{table}

\subsection{Experiment manifest}

C2 and C3 are batch properties, so no single transaction receipt can establish them. Every receipt references an experiment manifest containing: the ordered profile and condition sets; the complete expected assignment matrix; replication and retry policies; treatment flags; an allowlist of fields permitted to vary across arms; protocol fingerprints for prompts, schemas, parsers, normalizers, enforcers, choosers, scorers, model assignments, and interaction horizon; and the analysis contract, including estimand, nesting, pairing, estimator, threshold, seed, draw count, input order, and interval convention.

When the signed manifest predates committed ingress, it acts as a cryptographic preregistration. E1--E6 are explicitly retrospective imports: their manifests were reconstructed after the runs, so they support faithful mechanization rather than prospective confirmation. E7 is separate and prospective: its signed assignment matrix and OpenTimestamps proof precede all 30 ingress tickets.

\subsection{Transaction receipts and privacy}

The public receipt carries assignment identifiers, terminal status, manifest and artifact digests, a private-profile commitment, semantic digests for the trace, offer and outcome, exact claim values, and coverage counts. An auditor envelope opens the profile and preserves ordered raw-hidden and visible events, all attempts, raw and parsed offers, post-enforcement offers, chooser input/output, treatment events, reported scalars, and artifact references.

Low-entropy values such as WTP cannot safely use a bare hash. The specification commits a private canonical object $x$ as
\begin{equation}
\mathrm{HMAC\mbox{-}SHA256}_{n}\bigl(\texttt{tag}\,\|\,\mathrm{uint64be}(|x|)\,\|\,x\bigr),
\end{equation}
where $n$ is a fresh secret 32-byte nonce stored with the encrypted opening \cite{krawczyk1997hmac}. Canonical JSON follows RFC~8785 semantics \cite{rundgren2020jcs}; schema-declared sets may reorder, while transcript and attempt order may not. Economic amounts are scaled integers, aggregate means are $(\mathrm{sum},\mathrm{count})$, and decision-bearing interval endpoints retain a precision envelope. If a decision boundary intersects that envelope, the verifier abstains.

This architecture exposes a tension: verification needs access to the hidden WTP and profile used to score the experiment, while the marketplace policy may be designed to keep that information from the seller. The receipt therefore creates an auditor-visible encrypted evidence store, not a public transcript. CR-3 implements opening-key separation, but authorization, retention, and deletion remain governance obligations.

\subsection{Replay boundary and verdicts}

The replay claim has two tiers. For deterministic scripted sellers, replay begins at the committed policy artifact and continues through offer generation, normalization, buyer choice, and welfare. For LLM sellers, replay begins at the recorded raw output and continues through parsing, enforcement, choice, and scoring. We do not claim that another model call will reproduce the same text.

Before C1--C4, the verifier canonicalizes and deduplicates idempotent deliveries, checks receipt IDs and chain links, opens profile commitments, resolves assignment coverage, and retains retry order. It then recomputes all transaction claims exactly. C2 compares declared invariant fingerprints across arms. C1 and C3 average replications within profile cells before profile-level contrasts; generation rows never become independent buyers. C4 requires exact replay, $\mathrm{BS}+\mathrm{SP}=W$, and the full report bundle. A causal policy claim is licensed only when all four criteria pass and treatment fidelity is established. A descriptive claim may remain licensed when C1 or C3 is inconclusive but C2 and C4 pass.

\section{Implementation and Evaluation}

\subsection{Frozen specification and implementation}

We adversarially reviewed and froze ClaimReceipt v0.3 before implementing the verifier. Earlier v0.1 and v0.2 files and hashes remain immutable. Two non-verdict observations informed amendments: an integrality audit of 9,744 historical amount cells and inspection of the historical match-quality scorer. No C1--C4 verdict or fault result was observed before v0.3.
\begin{center}
\scriptsize SHA-256:\\
\texttt{18d10901d7760b1ab99fbeed97b2e37d}\\
\texttt{704092ba217f38478bd37d01378a1b81}
\end{center}

The reference implementation reaches CR-2 Batch Verification on the imported corpus and CR-3 Prospective Coverage on E7. CR-3 adds two Ed25519 signing roles, an X25519 auditor-opening role, pre-ingress assignment tickets and checkpoint, encrypted evidence, signed receipt chaining, and coverage-qualified output.

\subsection{Corpus}

We import 1,392 terminal records from the previously audited A2A study without importing its manual labels. The six retrospective epochs remain exactly those in Table~\ref{tab:corpus}. The deterministic epochs contain 600 records; the remaining 792 are LLM or fixed-menu records associated with LLM experiments. Buyer and seller generations use Qwen2.5-Instruct 1.5B, 3B, and 14B historically \cite{yang2024qwen25}. Retrospective receipts use deterministic test nonces. E7 is not added to the 1,392 count: it independently runs 15 profiles under NONE and BOTH with Qwen2.5-Instruct 3B, $k=1$, and fresh random nonces.

\begin{table}[t]
\caption{Retrospective corpus. Every epoch receives a separate manifest.}
\label{tab:corpus}
\centering
\scriptsize
\begin{tabularx}{\columnwidth}{@{}lrX@{}}
\toprule
Epoch & Receipts & Primary audit role \\
\midrule
E1 original scaffold & 270 & known C2 failure \\
E2 unified protocol & 450 & controlled treatment \\
E3 repeated subset & 36 & C3 stability \\
E4 incentive prompts & 36 & C1 manipulation \\
E5 scripted profit maximum & 300 & deterministic positive control \\
E6 scripted bundle stress & 300 & deterministic positive control \\
\midrule
Total & 1,392 & six manifests \\
\bottomrule
\end{tabularx}
\end{table}

\subsection{Evaluation questions}

R1 reproduces five manually labeled audit cells and reports the historical $4\times6$ grid; post-specification controls separately show C1/C3 can \pass{}. R2 measures exact replay. R3 compares evidence abstractions and removes 13 field groups. R4 applies frozen semantic faults and benign controls, including held-out retry shadowing versus a valid retained retry. R5 tests public versus authorized observability and omission propagation; R6 measures the prospective CR-3 path. Byte tampering remains an integrity sanity check rather than the headline.

\section{Results}

\subsection{R1: the verifier reproduces the manual audit}

The verifier agrees on all five manually labeled cells (5/5). It marks E1 C2 \invalid{} because the unguarded arm used a \texttt{room\_attributes/price} bundle and whole-bundle chooser while the guarded arm used a base-plus-add-ons schema and subset chooser. It marks E2 C2 \pass{} after those invariant fingerprints are unified. It returns \inconclusive{} for E4 C1 and E3 C3, and \pass{} for joint scripted accounting.

Table~\ref{tab:grid} gives the historical grid plus the separate E7 prospective row. No additional \invalid{} cell appears beyond E1 C2. We distinguish contract abstention ($I_C$: no required contract) from evidential abstention ($I_E$: the contract exists but evidence does not cross its boundary). C4 passes with authorized evidence even when another criterion blocks the causal comparison.

\begin{table}[t]
\caption{Verifier grid. P=\pass, X=\invalid; $I_C$ lacks a contract and $I_E$ has insufficient evidence. Cov is prospective coverage; retrospective rows are not applicable. A dagger marks the manually audited comparisons; E5--E6 form one joint accounting label.}
\label{tab:grid}
\centering
\small
\begin{tabular}{lccccc}
\toprule
Epoch & C1 & C2 & C3 & C4 & Cov \\
\midrule
E1 original & $I_C$ & X$^{\dagger}$ & $I_C$ & P & -- \\
E2 unified & $I_C$ & P$^{\dagger}$ & $I_C$ & P & -- \\
E3 repeated & $I_C$ & P & $I_E^{\dagger}$ & P & -- \\
E4 incentive & $I_E^{\dagger}$ & P & $I_C$ & P & -- \\
E5 scripted profit & $I_C$ & P & $I_C$ & P$^{\dagger}$ & -- \\
E6 scripted stress & $I_C$ & P & $I_C$ & P$^{\dagger}$ & -- \\
E7 prospective & $I_C$ & P & $I_C$ & P & P \\
\bottomrule
\end{tabular}
\end{table}

Agreement alone is evidence of faithful mechanization, not independent accuracy: a criterion-wise constant classifier could match five sparse labels. The verifier does not read those labels; it recomputes the underlying contrasts and intervals. Moreover, the two synthetic decision-rule controls return \pass{}: C1 has a monotone $+38.5$ contrast with one-sided 95\% lower bound 37.25, and C3 has a stable $+28.5$ contrast with central 95\% interval $[27.0,30.1]$. These post-specification controls rule out constant abstention but do not enter the 1,392-record corpus or support a behavioral claim.

The historical numerical evidence is likewise reproduced rather than copied. In E4, mean seller profit is 25.8 under the compliance prompt, 57.5 under the standard prompt, and 33.8 under profit pressure. The strongest-minus-weakest profile contrast is $+8.0$; its one-sided 95\% lower bootstrap bound is $-2.7$ (central 95\% interval $[-12.0,20.0]$), and the ordered response is non-monotone. In E3, the profile-level guardrail effect is $+37.6$ with 95\% interval $[-34.2,109.3]$ and mean within-cell generation SD 47.8. The verifier therefore abstains even though the point estimate is positive.

\subsection{R2: the replay boundary is exact}

All 600 deterministic records exactly match at policy, normalization, chooser, and metric stages. All 792 records in the LLM-associated epochs exactly match from the committed raw offer through choice and economic claims. Their recorded LLM text is hashed and auditable but not regenerated. The 210 fixed-menu rows inside those epochs additionally replay from menu policy. Thus the claim is strong where determinism exists and deliberately narrower where stochastic generation begins.

\begin{table}[t]
\caption{Exact replay by tier. Policy replay for the LLM-associated tier applies only to 210 deterministic fixed-menu rows; no LLM regeneration is claimed.}
\label{tab:replay}
\centering
\small
\begin{tabular}{lrrr}
\toprule
Replay tier & Policy & Choice & Metrics \\
\midrule
Scripted end-to-end & 600/600 & 600/600 & 600/600 \\
LLM post-generation & 210/210 & 792/792 & 792/792 \\
\bottomrule
\end{tabular}
\end{table}

\subsection{R3: generic evidence is not claim-sufficient}

Table~\ref{tab:abstraction} asks whether evidence uniquely determines a claim under Equation~\ref{eq:sufficiency}, not whether it stores a reported number. Outcome scalars, transcripts, and generic signed receipts determine none of the 12 claims because each omits required claim inputs, protocol, or batch evidence. An opened transaction receipt recovers six economic claims but not treatment fidelity or batch criteria; full ClaimReceipt determines all 12 within its declared scope.

\begin{table}[t]
\caption{Claims uniquely determined by each evidence abstraction. Counts include eight transaction/treatment claims and C1--C4.}
\label{tab:abstraction}
\centering
\scriptsize
\setlength{\tabcolsep}{2.5pt}
\begin{tabularx}{\columnwidth}{@{}Xr>{\raggedright\arraybackslash}X@{}}
\toprule
Evidence abstraction & Determined & Main missing distinction \\
\midrule
Outcome scalars & 0/12 & claim inputs and batch contract \\
Transcript + outcome & 0/12 & private state, chooser, protocol, coverage \\
Generic signed receipt & 0/12 & claim-specific semantics and artifacts \\
Opened transaction receipt & 6/12 & treatment fidelity and batch contract \\
ClaimReceipt v0.3 & 12/12 & none within declared claim class \\
\bottomrule
\end{tabularx}
\end{table}

Table~\ref{tab:dependencies} brings the frozen dependency matrix into the main paper. The ablation is an implementation-conformance check against this declared matrix, not a procedure that discovers dependencies from data. All 13 field-group removals make at least one claim underdetermined, confirming that the verifier does not silently trust a stored value after its declared evidence is removed. This establishes non-redundancy only under the tested group ablations, not global field minimality over all $2^n$ subsets.

\begin{table*}[t]
\caption{Frozen absolute field-to-claim dependencies. A dot means that removing the field group makes the claim underdetermined. Cmp=completion; FB=first-best fraction; MQ=match quality; IG/CG=information/conduct guardrail fidelity.}
\label{tab:dependencies}
\centering
\scriptsize
\setlength{\tabcolsep}{3.4pt}
\begin{tabular}{lcccccccccccc}
\toprule
Field & Cmp & BS & SP & W & FB & MQ & IG & CG & C1 & C2 & C3 & C4 \\
\midrule
I ingress/coverage & $\bullet$ & $\bullet$ & $\bullet$ & $\bullet$ & $\bullet$ & $\bullet$ & & & $\bullet$ & $\bullet$ & $\bullet$ & $\bullet$ \\
A assignment & & & & & & & & & $\bullet$ & $\bullet$ & $\bullet$ & $\bullet$ \\
P private profile & $\bullet$ & $\bullet$ & $\bullet$ & $\bullet$ & $\bullet$ & $\bullet$ & & $\bullet$ & $\bullet$ & & $\bullet$ & $\bullet$ \\
K catalog/cost & & & $\bullet$ & $\bullet$ & $\bullet$ & $\bullet$ & & & $\bullet$ & & $\bullet$ & $\bullet$ \\
T ordered trace & & & & & & & $\bullet$ & & & & & \\
O offer stages & $\bullet$ & $\bullet$ & $\bullet$ & $\bullet$ & $\bullet$ & $\bullet$ & & $\bullet$ & $\bullet$ & & $\bullet$ & $\bullet$ \\
D chooser & $\bullet$ & $\bullet$ & $\bullet$ & $\bullet$ & $\bullet$ & $\bullet$ & & $\bullet$ & $\bullet$ & & $\bullet$ & $\bullet$ \\
F scoring artifacts & & $\bullet$ & $\bullet$ & $\bullet$ & $\bullet$ & $\bullet$ & & & $\bullet$ & & $\bullet$ & $\bullet$ \\
G guardrail events & & & & & & & $\bullet$ & $\bullet$ & & $\bullet$ & & \\
Q protocol & & & & & & & $\bullet$ & $\bullet$ & $\bullet$ & $\bullet$ & $\bullet$ & $\bullet$ \\
J objective contract & & & & & & & & & $\bullet$ & & & \\
R replication contract & & & & & & & & & $\bullet$ & & $\bullet$ & $\bullet$ \\
B report bundle & & & & & & & & & & & & $\bullet$ \\
\bottomrule
\end{tabular}
\end{table*}

\subsection{R4: semantic faults and benign controls}

The verifier returns the frozen expected result on 11/11 semantic faults and raises no false positive on 0/8 benign variations. Detected cases include cross-arm schema drift, profile-opening conflict, a recorded guardrail flag without an enforcement event, scorer/version drift, pseudo-replication, selective reporting, missing terminal evidence, a missing private opening, and an incomplete report bundle. Most importantly, the held-out retry-shadowing fault returns C3 \invalid{}, while the paired benign retained retry remains \pass{}. Equivalent JSON key order, whitespace, integral numeric spelling, Unicode NFC, set order, idempotent redelivery, and transport-only timestamps do not alter scientific verdicts.

These rates are specification-conformance results on a small frozen suite, not estimates of performance against all possible attacks. L1 byte mutation is detected by standard digest checks. L4 interactions outside committed ingress remain unobservable and are reported as residual risk, not counted as false negatives.

\noindent\emph{External specification-legibility probe.}
Under the frozen single-response protocol the model exhausted its output budget before emitting final JSON; no valid measurement was obtained and the run is not scored. The frozen protocol under-provisioned output for a single-response format. A subsequent per-item run---explicitly post-hoc and not protocol-conformant---yielded 10/11 parseable responses, of which 3/11 matched the verifier. We treat this as exploratory diagnosis, not a confirmatory measurement.

The dominant error mode was temporal: the specification classifies faults by layer but never declares fault timing relative to commitment as an explicit attribute, although every layer assignment depends on it. The probe therefore motivates a v0.4 amendment adding \texttt{fault\_timing} $\in \{$\texttt{pre\_commitment}, \texttt{post\_commitment}$\}$ to each taxonomy entry, with a preflight rule routing post-commitment cases to integrity checking before layer classification. We state the amendment but do not adopt it: applying it after observing the probe would void the preregistration, and re-testing against the same eleven items would evaluate a specification written to those items.

This is one model under one protocol; we cannot separate specification ambiguity from model-specific reading failure, so the result bounds legibility from below rather than measuring it.

\subsection{R5--R6: prospective coverage, observability, and overhead}

E7 commits 15 profiles under NONE and BOTH before inference (internal IDs \texttt{unguarded\_agent} and \texttt{guardrailed\_agent}). The signature-bound manifest time is 08:13:17.826Z; calendar attestations are obtained at 08:13:20.192Z; the first of 30 preassigned ingress tickets is issued at 08:13:20.197Z. All receipts form one verified chain and reconcile to 24 accepted and six rejected terminals. The supplemental OpenTimestamps proof was upgraded to a Bitcoin block-header attestation at height 963833. We do not claim that this host independently validated the block: it has no configured Bitcoin Core node. Role separation is therefore real at the key level, not at the organizational-principal level.

Table~\ref{tab:cr3} shows two selective degradations. Removing the last terminal receipt leaves a valid 29-receipt prefix but returns \texttt{INCONCLUSIVE\_COVERAGE}; C4 and descriptive licensing fall with it, while C2 remains \pass{}. Propagating unresolved coverage to C4 is an implementation reading of complete accounting, not a new frozen decision threshold. Before the private-opening ablation, we freeze a prediction (SHA-256 \texttt{72c554...4c0f}): public receipts should retain coverage and C2, but all six economic and two treatment-fidelity claims should become \inconclusive{}. The observed result matches exactly. In particular, public SP is not trusted because its frozen dependency includes the private profile needed to recompute the completion branch.

\begin{table*}[t]
\caption{CR-3 results. P=\pass, I=\inconclusive, D=determined. Percentages use 2,000.14\,ms model inference per transaction.}
\label{tab:cr3}
\centering
\scriptsize
\begin{minipage}[t]{0.28\textwidth}
(a) Missing-terminal propagation\\[3pt]
\centering
\setlength{\tabcolsep}{3pt}
\begin{tabular}{lcc}
\toprule
Output & 30/30 & 29/30 \\
\midrule
Coverage & P & I \\
C2 protocol & P & P \\
C4 accounting & P & I \\
Descriptive & licensed & blocked \\
\bottomrule
\end{tabular}
\end{minipage}\hfill
\begin{minipage}[t]{0.31\textwidth}
(b) Public versus auditor view\\[3pt]
\centering
\setlength{\tabcolsep}{2pt}
\begin{tabular}{lcc}
\toprule
Claim group & Public & Opened \\
\midrule
Coverage + C2 & P & P \\
Six economic & I & 6/6 D \\
Two fidelity & I & 2/2 D \\
C4 / descriptive & I / blocked & P / licensed \\
\bottomrule
\end{tabular}
\end{minipage}\hfill
\begin{minipage}[t]{0.37\textwidth}
(c) Per-transaction cost\\[3pt]
\centering
\setlength{\tabcolsep}{3pt}
\begin{tabular}{lrr}
\toprule
Path & Cost & Inference \\
\midrule
Terminal write & 361\,$\mu$s & 0.018\% \\
Including ticket & 428\,$\mu$s & 0.021\% \\
Public verify & 297\,$\mu$s & 0.015\% \\
Auditor verify & 734\,$\mu$s & 0.037\% \\
Cold auditor read & 1,040\,$\mu$s & 0.052\% \\
Storage & 9.90\,KB & -- \\
Epoch verdict & 31.74\,ms & -- \\
\bottomrule
\end{tabular}
\end{minipage}
\end{table*}

The primary auditor-verification number is the 10-run median, 734\,$\mu$s (range 724--828); 1,040\,$\mu$s is one cold read, not a second estimator. Storage comprises a 2.64\,KB public receipt, 6.21\,KB encrypted envelope, and 1.05\,KB ingress ticket. Tickets are batch-issued after the epoch assignment set is committed and before inference, making their signing cost pre-ingress rather than terminal-path latency.

\section{Discussion and Limitations}

\noindent\emph{The contribution is semantics, not cryptography.}
A hash chain answers whether retained bytes changed; ClaimReceipt answers which retained distinctions a declared claim depends on. The field--claim matrix is the constructive dual of an identifiability failure: if removing a field group allows two admissible executions with different claim truth to share one evidence projection, no later solver or metric can repair that loss. Other domains will need different claim classes and dependencies, but the workflow---declare claims, derive evidence, commit the experiment, replay or abstain---transfers.

\noindent\emph{Selective abstention is a result.}
The grid contains many \inconclusive{} cells because historical runs were not designed for every criterion. This is intended behavior. The verifier still licenses exact descriptive accounting where C2 and C4 pass while blocking causal claims about strategic sellers or stable policy effects. A gate that never stops an attractive result is not verification.

\noindent\emph{Receipts do not create external coverage.}
E7 demonstrates prospective manifest, tickets, checkpoint, encrypted openings, chaining, and coverage-qualified \pass{}. The missing-terminal arm shows why this layer is not implied by replay. Yet CR-3 proves completeness only relative to committed ingress: a platform that prevents a transaction from reaching the witness remains outside observability (L4). Every \pass{} therefore carries \texttt{verified\_only\_within\_committed\_ingress}.

\noindent\emph{Privacy remains an engineering and governance burden.}
Auditable welfare requires the evaluator's hidden profile, including sensitive low-entropy values. E7 encrypts each opening to a separate X25519 role, but one operator still holds every role key. A deployment needs independent custody, authorization, retention, and deletion; otherwise verifiability can undermine the information protection being evaluated. Zero-knowledge metrics remain future work.

\noindent\emph{Evaluation scope.}
The evidence comes from one configurable hotel testbed and one principal model family; E7 uses only Qwen2.5-Instruct 3B and $k=1$. The 11 faults are designed rather than organic, and the abstraction analysis is scoped to the frozen claim class. We have not tested a second domain, an independent implementation, or separately operated trust roles. The next systems question is governance: independent ingress witnessing, key custody, and admission rules that reduce L4. The legibility probe also suggests a check others can apply cheaply: before treating a frozen contract as binding on third parties, measure whether an independent reader can predict its verdicts.

\section{Related Work}

\noindent\emph{Agent evaluation and observability.}
Agent evaluations emphasize interactive completion and scenario-aware measurement \cite{liu2023agentbench,kapoor2025agents,liang2022helm}, while trace audits show that aggregate success can conceal failures and misrank repairs \cite{zhu2026trace,zhu2026aggregate}. ClaimReceipt specifies evidence needed to recompute bounded claims and validity verdicts.

\noindent\emph{Provenance and tamper evidence.}
Timestamping, authenticated logs, and Certificate Transparency establish append-only integrity \cite{haber1991timestamp,crosby2009efficient,laurie2013certificate}; provenance tracks artifact dependencies \cite{moreau2013prov,boettiger2015docker}. We add claim semantics and conditional coverage using standard primitives unchanged.

\noindent\emph{Evaluation contracts and construct validity.}
Construct validity asks whether an operationalization measures its named object \cite{cronbach1955construct,jacobs2021measurement}; datasheets and model cards expose assumptions \cite{gebru2021datasheets,mitchell2019modelcards}. ClaimReceipt turns a subset into machine-checkable contracts that stop missing evidence from silently becoming a positive conclusion.

\section{Conclusion}

Agent verification needs both a sufficient record for each claim and a committed set against which omissions are visible. On 1,392 historical records, CR-2 reproduces a manual audit, replays deterministic and post-generation claims, and separates semantic faults from benign variation. On 30 prospective transactions, CR-3 distinguishes complete evidence, a missing terminal, and unavailable private openings at negligible runtime cost. The result is not a universal receipt, but a demonstrated method for deciding what agent-evaluation evidence licenses---including when it licenses only abstention.

\bibliographystyle{plain}
\bibliography{references}

\end{document}